\documentclass[letterpaper, 10 pt, conference]{ieeeconf}

\IEEEoverridecommandlockouts
\usepackage[T1]{fontenc}
\usepackage{mathptmx}   
\usepackage{booktabs}
\usepackage{graphicx}
\usepackage{wrapfig}
\usepackage[table]{xcolor}
\usepackage{amssymb}
\usepackage{amsmath}
\usepackage{multirow}
\usepackage{cellspace}
\usepackage{caption}
\usepackage{subcaption}
\usepackage{cite}
\usepackage[ruled,vlined,linesnumbered]{algorithm2e}

\title{\LARGE \bf
SWAP: Stepwise Action Policy Routing \\ for Vision-Language-Action Models
}

\author{Mousumi Das$^{*}$, Aditeya Prajapati$^{*}$, Abrar Anwar$^{\dagger}$, Jesse Thomason$^{\dagger}$ \\
{University of Southern California}
\thanks{$^*$ Equal Contribution. $\dagger$ Equal Advising. Contact: \texttt{\{mousumid,} \texttt{aditeyak,} \texttt{abrar.anwar,} \texttt{jessetho\}@usc.edu}, }
}
\begin{document}

\maketitle
\thispagestyle{empty}
\pagestyle{empty}


\begin{abstract}

Robot manipulation systems using Vision-Language-Action (VLA) model backbones typically use just one VLA for task execution. 
However, individual VLAs do not perform well across different task states and environments. 
We introduce a framework for dynamically composing multiple VLA policies during execution: $\text{\textbf{S}tep\textbf{W}ise \textbf{A}ction \textbf{P}olicy Routing (SWAP)}$. 
SWAP formulates policy routing as an offline reinforcement learning problem, learning a routing critic that selects the most appropriate policy at each decision step given the current observation. 
SWAP enables robots to select new policies to execute online rather than committing to a single policy for the duration of an episode. 
We evaluate SWAP on both real-world DROID manipulation tasks and LIBERO simulation experiments. 
SWAP improves over fixed-policy execution and routing baselines, giving absolute improvements in real-world task success up to 33\% while reducing successful trajectory robot action step length by 28.3\%. 
\end{abstract}



\section{Introduction}
Robot manipulation systems are increasingly powered by large foundation models such as Vision-Language-Action (VLA) models that directly map visual observations and language instructions to robot actions. 
These systems have demonstrated impressive generalization across manipulation tasks, but they are typically deployed under a restrictive assumption: that a single policy should control the robot for the duration of execution. 
In practice, however, no individual policy performs consistently well across every episode.
A policy that performs well under one scene configuration may fail under another, while a stronger policy overall may still make poor decisions in specific states. 
For robots operating in dynamic environments, effective execution may therefore require not a single policy, but a collection of complementary policies that can be selectively composed during execution.

Existing approaches largely ignore this setting by training a single large generalist VLA. 
Recent progress in scaling VLA models has focused on improving the capability of individual policies through larger datasets~\cite{droid, open_x_embodiment_rt_x_2023}, better architectures~\cite{kim2024openvla, octo}, broader pretraining~\cite{black2024pi0,bjorck2025gr00t}, and better post-training~\cite{physicalintelligence2025pi0.5}, and has shown increasingly capable general-purpose manipulation systems. 
However, these approaches implicitly assume that a single policy can robustly handle the full distribution of task states encountered during deployment. 
This assumption is particularly limiting in real-world manipulation, where distribution shift, compounding execution errors, and recovery behaviors can substantially alter the execution over time. 
Even when multiple policies are available, selecting one at the start of an episode commits the robot to a fixed behavior regardless of how execution evolves.

In this work, we instead formulate a \emph{policy routing} problem for robot manipulation. 
Rather than selecting a single policy for an entire episode, we consider the setting where a robot has access to multiple candidate policies and must dynamically determine which policy should act at each decision step. 
This introduces a new challenge: routing decisions must account for long-horizon downstream results, since a chosen locally-reasonable policy may ultimately lead to failure, while an alternative choice may enable recovery or successful completion. 
To address this, we introduce \textbf{S}tep-\textbf{W}ise \textbf{A}ction \textbf{P}olicy Routing (SWAP), a framework that formulates policy routing as an offline reinforcement learning problem over a discrete action space of policy identities.

SWAP learns a routing critic from offline robot interaction data containing successful and failed executions from multiple candidate policies. 
At deployment, the routing critic evaluates the current observation and dynamically selects the policy predicted to maximize task success. 
This enables the robot to compose complementary behaviors across policies within a single trajectory.

We evaluate SWAP in both real-world and simulated manipulation settings. 
On a real-world DROID~\cite{droid} setup, SWAP improves task success from 63.3\% for the best standalone policy to 96.7\%, while reducing successful trajectory length by 28.3\%. 
Across large-scale LIBERO simulation experiments, we further show that learned policy routing is effective.
Together, these results suggest that robust robot execution may emerge not solely from building stronger individual policies, but from learning how to effectively route between them.

\begin{figure*}[t]
    \centering
    \includegraphics[width=.8\linewidth]{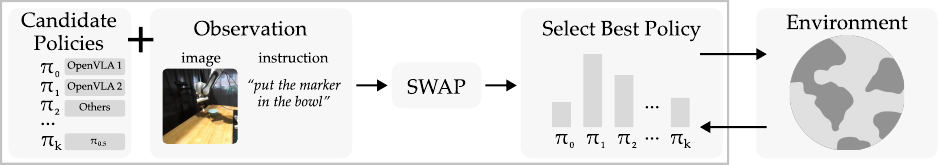}
    \caption{\textbf{Policy routing for robot manipulation.} 
    We introduce SWAP, an offline reinforcement learning framework for online policy routing.
    Given candidate robot policies, SWAP dynamically selects the best policy online for an instruction and current observation during task execution.}
        \label{fig:teaser}
\end{figure*}


\section{Background and Related Work}
\label{sec:related_work}
Recent progress in robot foundation models has produced a growing collection of policies with diverse capabilities and failure modes.
Our work draws connections between VLA models, routing methods in language models, and value-function-based decision making in robotics.

\textbf{Vision-Language-Action (VLA) Models} have demonstrated strong generalization across manipulation tasks by directly mapping visual observations and language instructions to robot actions~\cite{brohan2022rt1, kim2024openvla, black2024pi0, physicalintelligence2025pi0.5, octo}.
Work on evaluating robot policies has found that VLAs often remain brittle on certain types of distribution shifts such as language rephrasing, lighting changes, or increased clutter~\cite{wilbert_colloseum, simpler_env, anwar2024contrast, anwar2025efficient, kress2024robot, parekh2024investigating,xie2024decomposing,taxonomy2025arxiv,mitra2025roboticsteering, fei25libero}.
Finetuning VLAs on a new target task to mitigate these shifts can reduce robustness on the original training distribution~\cite{mitra2025roboticsteering}.
Our key insight relies on leveraging the diverse behavior of existing VLAs during a single task rollout.

\textbf{Routing in LLMs}
Recent work in large language models has explored routing mechanisms that dynamically select among multiple models, experts, or inference strategies at test time. 
Rather than relying on a single model for all inputs, these approaches allocate queries to specialized experts based on the current context~\cite{jiang2023llm,feng2025graphrouter,jitkrittum2025universal}, estimated difficulty~\cite{ding2025best,zhuang2025embedllm}, or predicted downstream performance~\cite{jiang2023llm, ong2024routellm, lu2024routing, shnitzer2023large,zhang2026dips}. 
Routing has been studied with model cascades~\cite{shen2025sater,yue2024large,moslem2026dynamic} and multi-model systems~\cite{aggarwal2023automix,ding2024hybrid,ong2024routellm,jiang2023llm} that balance cost factors~\cite{chen2023frugalgpt,ding2024hybrid} such as latency and capability by selecting the most appropriate model for a given request. 
A common theme across this literature is that different models exhibit complementary strengths, and that adaptive routing can outperform any individual model alone. 
In this work, we propose an adjacent problem of policy routing for robotics, where the routing decision must be made online throughout robot motion, as the robot's actions directly influence future observations and task outcomes.
Policy routing is also related to mixture-of-experts architectures, which learn a gating network jointly with the experts; in contrast, SWAP treats the candidate policies as fixed black boxes and learns only the router, so new policies can be added without retraining any expert.

\textbf{Value Functions for Robotics}
Value functions play a central role in robot learning by estimating the long-horizon consequences of actions and policy decisions. 
Model-free offline RL algorithms~\cite{kumar2020conservative, nakamoto2023cal, kostrikov2022offline, fujimoto2019off, kumar2019stabilizing, nair2020awac, awr} typically require value functions to train a robot policy.
These value functions enable these policies to predict actions that maximize long-term rewards. 
Recent work builds more general value~\cite{physicalintelligence2025pi0.5, ghasemipour2025selfimproving} and reward~\cite{zhang2025rewind, liang2026robometer, lee2026roboreward} functions.
Beyond policy optimization, value functions have been used to guide high-level decision making and planning. 
SayCan combined language model outputs with learned value functions to select skills that are both relevant and executable~\cite{ahn2022can}. 
Value functions have been used for failure detection~\cite{gokmen2023asking}, and more recently have been used to steer policies of larger VLAs at deployment time~\cite{nakamoto2024steering}.
We take inspiration from these works leveraging value functions at test-time, and focus on learning a routing critic that enables selection among multiple candidate robot policies for policy routing.


\section{Robot Policy Routing}
\label{sec:related_work}

While modern VLA policies can exhibit strong generalization, no single policy is uniformly optimal across all task states, environments, or execution regimes. 
A policy that performs well during coarse free-space motion may fail during contact-rich manipulation, while another may excel at precise object interaction yet behave conservatively during exploration or recovery. 
More broadly, policies trained with different data distributions, architectures, or objectives may exhibit complementary strengths and failure modes, motivating \emph{policy routing}.
Rather than committing to a single policy for an entire episode, we argue an agent should dynamically select which policy to execute next based on the current observation context, enabling multiple policies during a single rollout.

We formalize action taking in this setting as a Markov decision process (MDP), ${\mathcal{M} = (\mathcal{S}, \mathcal{A}, P, r, \gamma)}$, where at timestep $t$, the robot observes state $s_t \in \mathcal{S}$ and executes action $a_t \in \mathcal{A}$. Rather than learning a single policy $\pi(a_t \mid s_t)$, we assume access to a finite library of candidate robot policies,
${\mathcal{P} = \{\pi_1, \pi_2, \ldots, \pi_K\}},$
where each $\pi_k(a_t \mid s_t)$ represents a pretrained robot controller. In our setting, these policies may correspond to distinct VLA models trained with different architectures, datasets, or algorithms. 

Our key insight is to leverage the assumptions that each policy induces different behavior under identical observations and that no single $\pi_k$ dominates across all reachable states.
We aim to learn a \emph{routing policy}, $\phi$, that selects which robot policy to execute at each timestep. 
The routing policy is defined as $\phi(z_t \mid s_t),$ where $z_t \in \{1, \ldots, K\}$ denotes the selected policy index at time $t$. 
The robot action to take is then sampled from the chosen robot policy: $a_t \sim \pi_{z_t}(\cdot \mid s_t).$
We pose an objective to maximize long-horizon task performance for the routing policy: ${
    \text{argmax}_{\phi} \; \mathbb{E}_{\phi, P}\left[\sum_{t=1}^{T} \gamma^{t-1} r(s_t, a_t)\right].}$

This formulation differs from standard policy learning in that the action space of the router $\phi$ is not low-level motor control, but policy selection itself. 
The router operates at a higher level of abstraction, deciding \emph{which controller should act} rather than directly predicting robot actions.

\begin{figure*}[t]
    \centering
    \includegraphics[width=.9\linewidth]{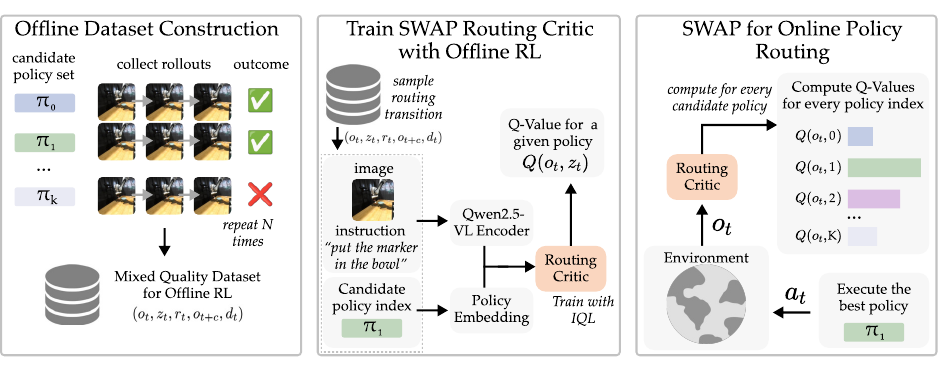}
    \vspace{-12pt}
\caption{\textbf{Overview of SWAP.}
(a) We construct an offline routing dataset by collecting trajectories from a set of candidate robot policies and chunking them into routing transitions labeled with sparse rewards and termination signals.
(b) SWAP trains a routing critic offline using IQL, where visual observations and task instructions are encoded with a frozen vision-language model and combined with a candidate policy embedding to predict policy-specific Q-values.
(c) SWAP evaluates all candidate policies under the current observation, selects the highest-valued policy, executes it in the environment for a short horizon, and repeats this process online throughout task execution.}
\label{fig:overview}
\end{figure*}

\section{SWAP: Stepwise Action Policy Routing}

We instantiate the policy routing framework as \textbf{S}tep-\textbf{W}ise \textbf{A}ction \textbf{P}olicy Routing (SWAP), an offline reinforcement learning approach for selecting among candidate robot policies during robot action rollouts (Figure~\ref{fig:overview}).
SWAP learns a routing critic that estimates which policy is most likely to lead to successful task completion given the robot's current execution state. 
Training is performed entirely offline from logged robot interaction data from pretrained VLAs, after which the learned critic is used at deployment to dynamically swap between VLA policies during execution.
The central component of SWAP is a routing critic that estimates the long-horizon utility of selecting a candidate robot policy given the current observation.
Given an observation $o_t$, the critic predicts the expected future returns of executing robot policies $\pi_i\in\mathcal{P}$ from that state onward.

\textbf{Offline Dataset Construction.}
To train the routing critic, we construct an offline dataset from robot interaction trajectories collected using each candidate policy across the target manipulation tasks that includes 
both successful and failed executions across all policies. 

Robot policies often predict and execute $c$ chunked low-level control actions. 
Rather than performing routing at every low-level control step, SWAP routes after each chunk as produced by the underlying VLAs.
Routing at the chunk level aligns routing decisions with the VLAs' own inference cadence, so no additional policy queries are introduced, and it avoids interrupting a policy mid-chunk, where intermediate actions are not individually meaningful.
Each routing transition therefore corresponds to executing a selected policy for a fixed chunk of robot actions before re-evaluating which policy should act next.
The resulting offline routing dataset consists of transitions $(o_t, z_t, r_t, o_{t+c}, d_t)$, where $o_t$ is the robot observation at the beginning of the chunk, $z_t$ denotes the selected policy, $r_t$ is the resulting sparse reward, $o_{t+c}$ is the observation after chunk execution, and $d_t$ indicates whether the episode terminated.
The reward signal is sparse and binary, reflecting whether successful task completion occurs following selection of a policy and the execution of its policy chunk.
This signal is particularly important in our routing setting, where the quality of a policy choice may only become apparent several timesteps later.

\textbf{Offline Critic Learning.}
Any offline reinforcement learning algorithm could be used to fit the critic given this offline routing dataset.
In our setting, the critic must accurately rank competing policy choices under mixed-quality offline data collected from multiple robot policies. 
We therefore instantiate SWAP using Implicit Q-Learning (IQL)~\cite{kostrikov2022offline}, which is well-suited for stable offline value learning.


\begin{figure*}[t]
    \centering
    \includegraphics[width=.8\linewidth]{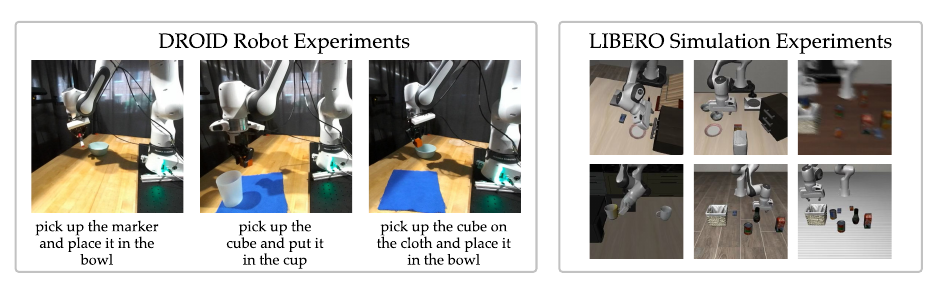}
    \vspace{-4pt}
\caption{\textbf{Evaluation environments.} We evaluate SWAP on both real-world DROID~\cite{droid} robot manipulation tasks and LIBERO~\cite{liu2023libero} simulation tasks. For simulation, we use the spatial subset of LIBERO-Plus~\cite{fei25libero}, which includes challenging out-of-distribution task variations.}   
\label{fig:datasets}
\end{figure*}

Given routing transitions $(o_t, z_t, r_t, o_{t+c}, d_t)$, IQL learns two routing critics, $Q_1$ and $Q_2$, along with a state-value function, $V$, by minimizing a temporal-difference objective over the two critics:
\begin{equation}
\begin{split}
\mathcal{L}_{Q} = \mathbb{E}_{(o_t,z_t,r_t,o_{t+c},d_t) \sim \mathcal{D}}
\Bigg[ \sum_{i=1}^{2} \Big( & Q_i(o_t, z_t) \\
& - \big(r_t + \gamma (1-d_t)V(o_{t+c})\big) \Big)^2 \Bigg].
\end{split}
\label{eq:iql}
\end{equation}
The value function is optimized separately via expectile regression over the learned Q-values. 
During both training and inference, SWAP uses the minimum of the two critics to produce conservative value estimates and reduce overestimation bias. 
Intuitively, offline RL encourages the critic to assign higher values to policy selections associated with successful downstream outcomes and lower values to those observed in failed trajectories. 
Since candidate policies may exhibit different strengths, failure modes, and robustness characteristics, continuously re-evaluating the values allows SWAP to select the policy best suited to the current  context rather than committing to a fixed controller for the duration of execution.
This may also encourage selecting policies that make faster downstream progress towards completing a task, as this critic is optimizing for the expected discounted return.
At inference time, the learned routing policy $\phi$ is induced directly from the critic by selecting the highest-valued candidate policy under the current observation.



\textbf{Routing Critic Architecture.}
We instantiate the routing critic using frozen visual-language embeddings from the Qwen2.5-VL-3B~\cite{qwen2.5-VL} VLM.
Using a frozen vision-language encoder allows SWAP to leverage strong pretrained visual and language representations while learning routing behavior from relatively small offline robot datasets.
The VLM encodes the current  visual observation and language instruction into a 2048-dimensional embedding.
This embedding is concatenated with a learnable embedding corresponding to the candidate policy identity and passed into a lightweight double-Q critic architecture consisting of MLPs that predict scalar action-values over the candidate policy set, along with a separate value network used during IQL training.


\textbf{Test-Time Policy Routing.}
Once the routing critic has been trained, SWAP uses it at deployment to dynamically select among candidate robot policies throughout task execution.
Given the current robot observation $o_t$, SWAP evaluates the routing critic over the candidate policy set, and selects the policy with the highest predicted long-horizon utility ${z_t =  \arg\max_{z} \min(Q_1(o_t,z), Q_2(o_t,z))}$.
The selected policy is queried to produce a short-horizon chunk of robot actions to execute.
This routing procedure repeats after each low-level chunk of robot actions is executed, allowing SWAP to dynamically swap between policies as execution unfolds.


\begin{algorithm}[b]
\small
\caption{High-level SWAP Algorithm}
\label{alg:swap}
\DontPrintSemicolon
\KwIn{Policies $\mathcal{P}=\{\pi_1,\dots,\pi_K\}$; VLM $f$; chunk size $c$}
\KwOut{Routing critics $Q_1, Q_2$}
\tcp{Offline dataset construction}
\ForEach{$\pi_k \in \mathcal{P}$, task}{
  Roll out $\pi_k$; store transitions $(o_t, z_t, r_t, o_{t+c}, d_t)$ in $\mathcal{D}$\;
}
\tcp{Offline critic learning via IQL}
\While{not converged}{
  Sample batch from $\mathcal{D}$; \ $h \gets f(o, \text{instruction})$\;
  Update $Q_1, Q_2$ with $\mathcal{L}_Q$ (Eq.~\ref{eq:iql}); update $V$\;
}
\tcp{Online stepwise routing}
\While{not done}{
  $h_t \gets f(o_t, \text{instruction})$\;
  $z_t \gets \arg\max_k \min\big(Q_1(h_t,k),\, Q_2(h_t,k)\big)$\;
  Execute $\pi_{z_t}$ for $c$ actions; \ $t \gets t + c$\;
}
\end{algorithm}


\section{Experiments and Results}
\label{sec:results}



We evaluate SWAP in both real-world and simulated manipulation settings to assess whether learned stepwise policy routing improves robot manipulation performance over fixed policy execution. In each setting, SWAP is trained using offline interaction data collected from the candidate policies. 


We compare SWAP against two classes of baselines. First, we evaluate each candidate policy independently, corresponding to the standard setting where a single policy is executed for the full episode without routing. 
Second, we compare against random routing, where candidate policies are selected uniformly at each control step without a learned critic, to isolate the contribution of learned routing from the benefits of policy switching alone.





\begin{table*}[t]
\centering
\small

\caption{\textbf{Real Robot Policy Routing.} We present real-world policy routing results on a Franka DROID setup across three manipulation tasks.
SWAP achieves the strongest overall performance, improving success rate from 63.3\% for the best fixed policy baseline to 96.7\%, demonstrating that learned policy routing can effectively compose complementary behaviors across candidate policies. 
Additionally, among successful trajectories, SWAP completes tasks in 28.3\% fewer steps than the best-performing individual policy (193 vs. 269 steps).
}
\setlength{\tabcolsep}{4pt}
\begin{tabular}{lrr rr rr rr rr}
\toprule
& \multicolumn{6}{c}{\textbf{Individual Policy}}
& \multicolumn{4}{c}{\textbf{Routing}} \\
\cmidrule(lr){2-7}
\cmidrule(lr){8-11}
& \multicolumn{2}{c}{\textbf{$\pi_0$}} 
& \multicolumn{2}{c}{\textbf{$\pi_{0.5}$}} 
& \multicolumn{2}{c}{\textbf{$\pi_0$-FAST}} 
& \multicolumn{2}{c}{\textbf{Random}}
& \multicolumn{2}{c}{\textbf{SWAP (Ours)}} \\
\cmidrule(lr){2-3}
\cmidrule(lr){4-5}
\cmidrule(lr){6-7}
\cmidrule(lr){8-9}
\cmidrule(lr){10-11}
\textbf{Task} & \textbf{Succ.} & \textbf{Steps} 
& \textbf{Succ.} & \textbf{Steps}
& \textbf{Succ.} & \textbf{Steps}
& \textbf{Succ.} & \textbf{Steps}
& \textbf{Succ.} & \textbf{Steps} \\
\midrule
Marker $\rightarrow$ Bowl 
& 30\% & 274 
& 60\% & 249
& 80\% & 209
& 60\% & 233
& \textbf{100\%} & 200 \\

Cube $\rightarrow$ Cup 
& 50\% & 358
& 20\% & 420
& 40\% & 372
& 60\% & 336
& \textbf{90\%} & 224 \\

Cube on Cloth $\rightarrow$ Bowl 
& 40\% & 288
& 50\% & 201
& 70\% & 227
& 80\% & 262
& \textbf{100\%} & 155 \\

\midrule
\textbf{Average}
& {40\%} & {307}
& {43.3\%} & {290}
& {63.3\%} & {269}
& {66.7\%} & {277}
& \textbf{96.7\%} & \textbf{193} \\
\bottomrule
\end{tabular}

\label{tab:droid_results}
\end{table*}

\subsection{Experiment Setup}

\textbf{Real-World DROID.}
We evaluate SWAP on real-world tabletop manipulation using the DROID Franka setup. 
The platform consists of a 7-DoF Franka Panda arm with RGB observations from multiple external and wrist-mounted cameras. 
For these experiments, SWAP aggregates Q-values across multiple camera views during inference to improve robustness to viewpoint ambiguity and occlusion. 
We evaluate three manipulation tasks: placing a marker into a bowl, placing a cube into a cup, and transferring a cube from a cloth into a bowl. 
Our candidate policy set consists of three policies from Physical Intelligence: $\pi_0$~\cite{black2024pi0}, $\pi_{0.5}$~\cite{physicalintelligence2025pi0.5}, and $\pi_0$-FAST~\cite{pertsch2025fast}.
We use the publicly released DROID checkpoints of $\pi_{0.5}$ and $\pi_0$-FAST zero-shot; we fine-tune $\pi_0$ on a small amount of data since it did not present zero-shot capabilities on our tasks. 

\textbf{LIBERO Simulation.}
We additionally evaluate SWAP in simulation on the LIBERO~\cite{liu2023libero} and LIBERO-Plus~\cite{fei25libero} benchmarks, focusing specifically on the LIBERO-Spatial subset.
To further test the generalization of SWAP, we evaluate on the spatial subset of LIBERO-Plus.
This setting enables larger-scale evaluation across diverse manipulation tasks. 
This simulation enables us to vary the composition of the candidate policy set across experiments, using $\pi_0$~\cite{black2024pi0}, OpenVLA~\cite{kim2024openvla}, and OpenVLA-OFT~\cite{kim2025openvlaoft} variants, in order to evaluate routing behavior under different policy qualities and larger candidate policy sets.


\textbf{Training Procedure.}
To train SWAP, we collect offline interaction trajectories from each candidate policy across the evaluated tasks. 
For DROID, we collect 10 successful and 10 failed rollouts per policy per task, for a total of 180 trajectories across three tasks and three policies; each method is then evaluated on 10 held-out trials per task. 
In practice, this data can come from the evaluations of the base policies.
For LIBERO, we collect 50 rollouts per policy per task on 8 of the 10 LIBERO-Spatial tasks, holding out tasks 5 and 9 for evaluation. We also evaluate the more diverse LIBERO-Plus Spatial subset.
We use sparse rewards in the range where successful task completion sets a final transition reward of 10, failure of 0, and in both cases all intermediate transitions receive 0 reward. 
We deliberately use only sparse task-completion reward signals and do not explore intermediate reward shaping, as that would require denser rewards that may be difficult to acquire. 
We train separate routing critics for the DROID and LIBERO environments. 
For real-world experiments, we train for 4k--8k epochs; for simulation we train for 200 epochs.

\subsection{Results}
\label{sec:results}

\textbf{SWAP outperforms individual policies and random routing.} As shown in Table~\ref{tab:droid_results}, across all DROID tasks, SWAP achieves the strongest overall performance, reaching 96.7\% success compared to 63.3\% for the best fixed policy baseline. 
SWAP achieves a +33.4 point absolute improvement over the strongest individual policy, and substantially larger gains over weaker fixed policies. 
These results suggest that learned stepwise routing can effectively compose complementary behaviors from multiple candidate policies, and it yields more reliable success rates than committing to any single policy for the duration of a full episode.
We additionally compare against random routing to isolate the contribution of learned policy routing from the benefits of switching alone. 
Random routing slightly outperforms the best fixed policy on average (66.7\% vs.\ 63.3\%), which we attribute to switching acting as a crude recovery mechanism: a policy that has stalled or drifted is replaced within a single chunk, whereas a fixed policy remains stuck until timeout. 
This indicates that the ability to switch is itself valuable, but SWAP's further +30 point improvement over random routing shows that most of the gain comes from learning \emph{which} policy to switch to.

\begin{table}[t]
\vspace{-8pt}
\centering
\small
\setlength{\tabcolsep}{4pt}
\caption{\textbf{Policy preference shifts as execution progresses.}
Policy selection frequencies across normalized trajectory segments. 
SWAP begins with nearly equal use of both policies but concentrates on $\pi_{0.5}$ during later stages of execution. This suggests that SWAP discovers policy strengths and exploits them at different phases of task completion.}
\begin{tabular}{lcc}
\toprule
\textbf{Segment} & $\pi_0$-FAST & $\pi_{0.5}$ \\
\midrule
First Third  & 51.0\% & 49.0\% \\
Middle Third & 22.6\% & 77.4\% \\
Final Third  & 36.3\% & 63.7\% \\
\bottomrule
\end{tabular}
\label{tab:policy_occupancy}
\vspace{-10pt}
\end{table}

\begin{table*}[t]
\centering
\small
\setlength{\tabcolsep}{4pt}

\begin{minipage}[t]{0.55\textwidth}
\centering
\textbf{(a) LIBERO with five policies}\\
\begin{tabular}{lcc}
\toprule
\textbf{Method} & LIBERO & LIBERO-Plus \\
\midrule
$\pi_0$ & 81\% & 54.2\% \\
OpenVLA-Spatial & 62\% & 0\% \\
OpenVLA-OFT & 50.6\% & 69.0\% \\
OpenVLA-OFT$_{\text{no-wrist}}$ & 59.8\% & 63.5\% \\
OpenVLA-OFT+ & 54.8\% & 80.8\% \\
\midrule
Random & - & 2.7\% \\

SWAP & - & 62.65\% \\
SWAP w/o $\pi_0$ & - & 70.65\% \\
SWAP w/o OpenVLA-Spatial & - & 63.00\% \\
\bottomrule
\end{tabular}
\end{minipage}
\hfill
\begin{minipage}[t]{0.40\textwidth}
\centering
\textbf{(b) Dominant policy ablation in LIBERO-Plus}\\
\begin{tabular}{lcc}
\toprule
\textbf{Method} & LIBERO & LIBERO-Plus \\
\midrule
$\pi_0$ & 81\% & 49.4\% \\
$\pi_{0.5}$ & 95\% & 92.0\% \\
\midrule
Random & 90\% & 79.0\% \\
SWAP & 97.5\% & 89.0\% \\
\bottomrule
\end{tabular}
\end{minipage}


\vspace{2mm}
\caption{\textbf{LIBERO experiments.}
\textbf{(a)} Performance of five candidate policies and various routing formulations. 
\textbf{(b)} We construct a two-policy setting where $\pi_{0.5}$ is the clear best policy in the LIBERO-Plus subset. SWAP matches or slightly outperforms the performance of the dominant policy.
}
\label{tab:libero_exps}

\vspace{-4mm}
\end{table*}
\begin{table}[t]
\centering
\caption{\textbf{Real Robot Policy Routing under OOD Distractors.} We add unseen distractor objects and evaluate each method over 10 trials per task. SWAP maintains high success despite the distribution shift.}
\label{tab:droid_ood}
\resizebox{\columnwidth}{!}{%
\begin{tabular}{lcccc}
\toprule
& \multicolumn{3}{c}{Individual Policy} & Routing \\
\cmidrule(lr){2-4} \cmidrule(lr){5-5}
Task with Distractors & $\pi_0$ & $\pi_{0.5}$ & $\pi_0$-FAST & SWAP (Ours) \\
\midrule
Marker $\rightarrow$ Bowl        & 10\% & 40\% & 40\% & 90\% \\
Cube on Plate $\rightarrow$ Bowl & 10\% & 30\% & 0\% & 80\% \\
\midrule
Average                                          & 10\% & 35\% & 20\% & 85\% \\
\bottomrule
\end{tabular}%
}
\end{table}

\textbf{SWAP reduces trajectory length.} 
SWAP not only improves task success rate, but also substantially reduces the number of actions required for successful completion, averaging 193 steps compared to 269 steps for the best-performing individual policy, a 28.3\% reduction. 
This result suggests that SWAP removes unnecessary behaviors by selecting more appropriate policies throughout execution, rather than requiring recovery from suboptimal policy choices. 
This behavior may emerge as a consequence of the routing critic optimizing for expected discounted return, which implicitly favors policy selections that make consistent downstream progress toward task completion.

\begin{figure*}[t]
    \centering
    \includegraphics[width=.85\linewidth]{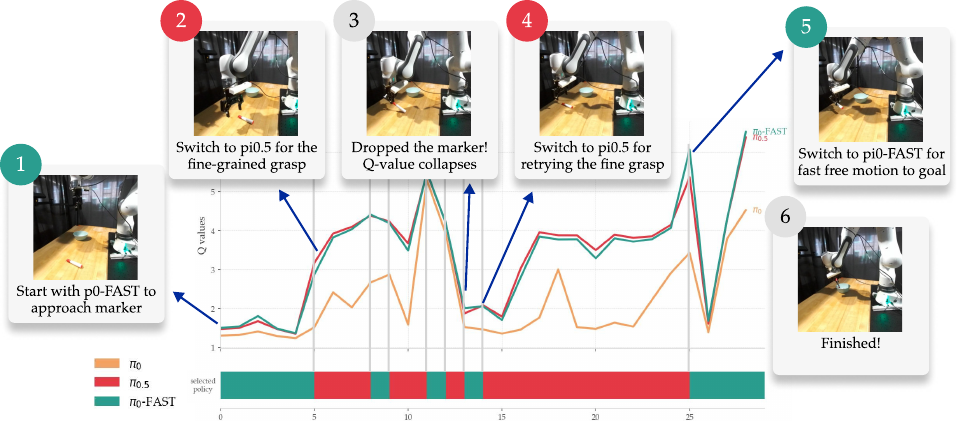}
    \caption{\textbf{Q-values and policy routing on the DROID marker-in-bowl task.} 
    SWAP scores $\pi_0$-FAST, $\pi_0$, and $\pi_{0.5}$ based on the predicted Q-values of each policy throughout task execution. 
    The images highlight key routing decisions in one SWAP rollout.
    The plot shows the corresponding Q-values over time, as well as the policy selected by SWAP at each decision step. 
    Initially, SWAP favors $\pi_0$-FAST, which takes larger steps during free-space movement. As the robot approaches the object, it switches to $\pi_{0.5}$ for the more precise grasp. 
    After the marker is accidentally dropped, the Q-values collapse and SWAP again selects $\pi_{0.5}$ to recover and reattempt the grasp. 
    Once the marker is secured, SWAP switches back to $\pi_0$-FAST for faster motion to the goal.
}    \label{fig:q_values}
\end{figure*}

\textbf{Q-value analysis.} Additionally, Figure~\ref{fig:q_values} visualizes how the learned Q-values evolve throughout execution. 
We observe that $\pi_0$ is consistently assigned low value and rarely selected, while $\pi_{0.5}$ and $\pi_0$-FAST regularly alternate as the preferred policy. 
When analyzing the learned Q-values, we find that the weaker standalone policy \(\pi_{0\text{-FAST}}\) has a lower value 36\% of the time, while (\(\pi_{0.5}\))  is assigned the higher value 64\% of the time.
This difference is reflected in the routing behavior itself, where SWAP changes the selected policy an average of 8.7 times per rollout, with each switch corresponding to an average Q-value improvement of +0.117 over the previously active policy.

\textbf{Policy routing evolves over trajectory.} We further analyze aggregate routing behavior in Table~\ref{tab:policy_occupancy} by measuring policy selection frequency across normalized trajectory segments. 
Routing is relatively balanced early in execution, but increasingly shifts toward $\pi_{0.5}$ as trajectories progress, with $\pi_{0.5}$ accounting for 77.4\% of selections in the middle third and 63.7\% in the final third, while $\pi_0$ is effectively never selected due to its consistently low predicted value. 
These results suggest that SWAP learns structured policy routing rather than randomly composing candidate policies.

\textbf{Qualitative analysis.}
To qualitatively reinforce these quantitative notions of how routing behavior changes over time, we also look into how SWAP learns to specialize on when to call different policies in Figure~\ref{fig:q_values}.
Early in the rollout, SWAP selects $\pi_{0\text{-FAST}}$ while the robot traverses free space toward the target object. As the robot approaches the marker and the task requires more precise manipulation, SWAP switches to $\pi_{0.5}$, which consistently receives higher Q-values during the grasping phase. After the marker is accidentally dropped, the predicted Q-values for all policies decrease as expected. SWAP then re-selects $\pi_{0.5}$ to recover and successfully reattempt the grasp. Once the marker has been secured, the Q-values again favor $\pi_{0\text{-FAST}}$, and SWAP switches back to it for faster movement to the bowl. 
These behaviors suggest that the learned routing critic captures complementary policy strengths, preferring $\pi_{0\text{-FAST}}$ for free-space motion and $\pi_{0.5}$ for fine-grained object manipulation and recovery. 
We found that for the individual policies, after a failure, there would be no attempts at recovery beyond simple hovering until a timeout was reached.
We did not observe visible discontinuities in robot motion when SWAP switches policies; since all candidate policies consume the same observation and output actions in the same DROID action space, switching amounts to querying a different model for the next chunk from the current state.

\textbf{Routing transfers under distribution shift.}
We evaluate SWAP under OOD conditions in both LIBERO and DROID settings. 
For these OOD experiments, we re-train all 3 base policies on the in-distribution tasks, and add unseen distractor objects to two DROID tasks and evaluate SWAP over 10 trials each without retraining the critic (Table~\ref{tab:droid_ood}), achieving 90\% and 80\% success, compared to 40\% and 30\% for the best individual policies. 
Similar to our qualitative analysis section, we found strong correlation of policy swapping during key stages of execution like grasping, moving, and dropoff. 
In simulation, Table~\ref{tab:libero_exps} (a) presents a setting with five heterogeneous policies whose ranking inverts under shift: $\pi_0$ is the strongest in-distribution (81\%) but only fourth on LIBERO-Plus (54.2\%). 
SWAP achieves 62.65\%, far above random routing (2.7\%), which collapses here because switching between behaviorally incompatible policies is destructive; it also correctly ignores OpenVLA-Spatial (0\%), whose removal changes performance by only 0.35 points. 
The remaining gap to the best fixed policy stems from $\pi_0$: removing it raises SWAP to 70.65\%. 
Since $\pi_0$ is the best policy on the critic's training distribution, this preference is correct for the data SWAP saw; the failure is that in-distribution quality does not predict OOD quality. 
The router itself thus generalizes to novel scenes, with the limiting factor being whether candidate strengths are stable across the shift.

\textbf{SWAP maintains performance when selecting between a dominant policy and a poorly performing policy.}
For Table~\ref{tab:libero_exps} (b), we design a scenario to evaluate SWAP in a setting where the candidate policy set contains one strong policy and one substantially weaker policy. 
Each of these policies is trained and tested on the in-distribution tasks of LIBERO, however, to test the generalization ability of SWAP, we only train SWAP on 8 of the 10 core tasks in LIBERO-Spatial. 
Thus, these evaluation results on LIBERO are on two tasks that are unseen to SWAP, while they are seen to the other methods.
Even so, SWAP matches or outperforms all methods, including the dominant policy.
In LIBERO-Plus, SWAP achieves an 89\% success rate, closely matching the stronger standalone policy despite the presence of a significantly weaker alternative, while random routing degrades to 79\% success. 
These results suggest that SWAP is able to effectively identify and prioritize stronger policy behaviors throughout execution rather than being negatively affected by lower-performing candidate policies.

\section{Conclusion}
\label{sec:conclusion}

We introduce SWAP, a framework for learned stepwise policy routing across Vision-Language-Action policies.
Rather than committing to a single policy for the duration of an episode, SWAP learns a routing critic that dynamically selects among candidate policies based on the robot's current observation.
Across both real-world DROID evaluations and large-scale LIBERO simulation experiments, we show that learned policy routing improves over fixed-policy execution and non-adaptive routing baselines.
These results suggest that individual robot policies often exhibit complementary strengths, and that explicitly routing between them can result in more reliable manipulation.
More broadly, SWAP highlights policy routing as a simple but effective mechanism for composing heterogeneous robot policies.

\textbf{Limitations.} 
Our experiments focus on routing across a small set of candidate policies, and it remains an open question how learned policy routing scales to dozens or hundreds of specialized models, especially under increasingly OOD scenarios.
Our evaluation is also restricted to short-horizon tabletop manipulation; on longer-horizon tasks the sparse terminal reward becomes a weaker training signal for the routing critic, and intermediate subtask completion signals would likely be needed.
As the candidate set grows, querying multiple heterogeneous policies also introduces systems challenges that may make on-device deployment intractable.
Finally, our formulation optimizes only for task return and does not account for costs such as inference latency or monetary cost. Incorporating cost-aware objectives while scaling routing to more heterogeneous policies is an exciting direction for future work.

\textbf{AI Tool Usage}: ChatGPT and Claude were used for proofreading the paper and as coding assistants.


\section*{Acknowledgments}
This work was supported in part by a DARPA ARC Safe and Assured Foundation Robots for Open Environments (SAFRON) grant (HR0011-25-3-0154), an Army Research Laboratory (ARL) Army Artificial Intelligence Innovation Institute (A2I2) grant (W911NF-23-2-0010), and by the NVIDIA Academic Grant Program using a DGX Spark.


\bibliographystyle{IEEEtran}
\bibliography{example}

@inproceedings{zhang2026dips,
  title={DiPS: Dialogue Policy Selection for High-Stakes Persuasion Agents},
  author={Zhang, Tianyi and Das, Mousumi and Anwar, Abrar and Thomason, Jesse and Traum, David},
  booktitle={Annual Meeting of the Special Interest Group on Discourse and Dialogue (SIGDIAL)},
  year={2026}
}

@misc{qwen2.5-VL,
    title = {Qwen2.5-VL},
    url = {https://qwenlm.github.io/blog/qwen2.5-vl/},
    author = {Qwen Team},
    month = {January},
    year = {2025}
}

@article{
lee2026roboreward,
title={RoboReward: General-Purpose Vision-Language Reward Models for Robotics},
author={Tony Lee and Andrew Wagenmaker and Karl Pertsch and Percy Liang and Sergey Levine and Chelsea Finn},
journal = {arXiv preprint arXiv:2601.00675},
year    = {2026},
}

@inproceedings{
ghasemipour2025selfimproving,
title={Self-Improving Embodied Foundation Models},
author={Seyed Kamyar Seyed Ghasemipour and Ayzaan Wahid and Jonathan Tompson and Pannag R Sanketi and Igor Mordatch},
booktitle={Advances in Neural Information Processing Systems (NeurIPS)},
year={2025},
}

@inproceedings{jiang2023llm,
  title={Llm-blender: Ensembling large language models with pairwise ranking and generative fusion},
  author={Jiang, Dongfu and Ren, Xiang and Lin, Bill Yuchen},
  booktitle={Annual Meeting of the Association for Computational Linguistics (ACL)},
  year={2023}
}

@inproceedings{shen2025sater,
  title={SATER: A Self-Aware and Token-Efficient Approach to Routing and Cascading},
  author={Shen, Yuanzhe and Liu, Yide and Huang, Zisu and Yin, Ruicheng and Zheng, Xiaoqing and Huang, Xuan-Jing},
  booktitle={Conference on Empirical Methods in Natural Language Processing (EMNLP)},
  year={2025}
}

@inproceedings{ding2024hybrid,
  title={Hybrid llm: Cost-efficient and quality-aware query routing},
  author={Ding, Dujian and Mallick, Ankur and Wang, Chi and Sim, Robert and Mukherjee, Subhabrata and R{\"u}hle, Victor and Lakshmanan, Laks and Awadallah, Ahmed H},
  booktitle={International Conference on Learning Representations (ICLR)},
  year={2024}
}

@article{jitkrittum2025universal,
  title={Universal model routing for efficient llm inference},
  author={Jitkrittum, Wittawat and Narasimhan, Harikrishna and Rawat, Ankit Singh and Juneja, Jeevesh and Wang, Congchao and Wang, Zifeng and Go, Alec and Lee, Chen-Yu and Shenoy, Pradeep and Panigrahy, Rina and others},
  journal={International Conference on Learning Representations (ICLR)},
  year={2026}
}

@inproceedings{feng2025graphrouter,
  title={Graphrouter: A graph-based router for llm selections},
  author={Feng, Tao and Shen, Yanzhen and You, Jiaxuan},
  booktitle={International Conference on Learning Representations (ICLR)},
  year={2025}
}

@inproceedings{zhuang2025embedllm,
  title={Embedllm: Learning compact representations of large language models},
  author={Zhuang, Richard and Wu, Tianhao and Wen, Zhaojin and Li, Andrew and Jiao, Jiantao and Ramchandran, Kannan},
  booktitle={International Conference on Learning Representations (ICLR)},
  year={2025}
}

@article{ding2025best,
  title={Best-route: Adaptive llm routing with test-time optimal compute},
  author={Ding, Dujian and Mallick, Ankur and Zhang, Shaokun and Wang, Chi and Madrigal, Daniel and Garcia, Mirian Del Carmen Hipolito and Xia, Menglin and Lakshmanan, Laks VS and Wu, Qingyun and R{\"u}hle, Victor},
  journal={International Conference on Machine Learning (ICML)},
  year={2025}
}

@article{moslem2026dynamic,
  title={Dynamic model routing and cascading for efficient LLM inference: A survey},
  author={Moslem, Yasmin and Kelleher, John D},
  journal={arXiv preprint arXiv:2603.04445},
  year={2026}
}

@inproceedings{yue2024large,
  title={Large language model cascades with mixture of thought representations for cost-efficient reasoning},
  author={Yue, Murong and Zhao, Jie and Zhang, Min and Du, Liang and Yao, Ziyu},
  booktitle={International Conference on Learning Representations},
  volume={2024},
  pages={21691--21728},
  year={2024}
}

@inproceedings{lu2024routing,
  title={Routing to the expert: Efficient reward-guided ensemble of large language models},
  author={Lu, Keming and Yuan, Hongyi and Lin, Runji and Lin, Junyang and Yuan, Zheng and Zhou, Chang and Zhou, Jingren},
  booktitle={Conference of the North American Chapter of the Association for Computational Linguistics (NAACL)},
  year={2024}
}

@article{shnitzer2023large,
  title={Large language model routing with benchmark datasets},
  author={Shnitzer, Tal and Ou, Anthony and Silva, M{\'\i}rian and Soule, Kate and Sun, Yuekai and Solomon, Justin and Thompson, Neil and Yurochkin, Mikhail},
  journal={International Conference on Learning Representations (ICLR)},
  year={2024}
}

@article{aggarwal2023automix,
  title={Automix: Automatically mixing language models},
  author={Aggarwal, Pranjal and Madaan, Aman and Anand, Ankit and Potharaju, Srividya Pranavi and Mishra, Swaroop and Zhou, Pei and Gupta, Aditya and Rajagopal, Dheeraj and Kappaganthu, Karthik and Yang, Yiming and others},
  journal={arXiv preprint arXiv:2310.12963},
  year={2023}
}

@article{chen2023frugalgpt,
title={FrugalGPT: How to Use Large Language Models While Reducing Cost and Improving Performance},
author={Chen, Lingjiao and Zaharia, Matei and Zou, James},
journal={Transactions on Machine Learning Research (TMLR)},
year={2023}
}

@article{ong2024routellm,
  title={Routellm: Learning to route llms with preference data},
  author={Ong, Isaac and Almahairi, Amjad and Wu, Vincent and Chiang, Wei-Lin and Wu, Tianhao and Gonzalez, Joseph E and Kadous, M Waleed and Stoica, Ion},
  journal={International Conference on Learning Representations (ICLR)},
  year={2025}
}

@inproceedings{gokmen2023asking,
  title={Asking for Help: Failure Prediction in Behavioral Cloning through Value Approximation},
  author={Gokmen, Cem and Ho, Daniel and Khansari, Mohi},
  booktitle={International Conference on Robotics and Automation (ICRA)},
  year={2023}
}

@article{liu2023libero,
  title={Libero: Benchmarking knowledge transfer for lifelong robot learning},
  author={Liu, Bo and Zhu, Yifeng and Gao, Chongkai and Feng, Yihao and Liu, Qiang and Zhu, Yuke and Stone, Peter},
  journal={Neural Information Processing Systems (NeurIPS)},
  year={2023}
}

@article{fei25libero,
    title={LIBERO-Plus: In-depth Robustness Analysis of Vision-Language-Action Models},
    author={Senyu Fei and Siyin Wang and Junhao Shi and Zihao Dai and Jikun Cai and Pengfang Qian and Li Ji and Xinzhe He and Shiduo Zhang and Zhaoye Fei and Jinlan Fu and Jingjing Gong and Xipeng Qiu},
    journal = {arXiv preprint arXiv:2510.13626},
    year={2025},
}

@article{ahn2022can,
title={{Do As I Can and Not As I Say: Grounding Language in Robotic Affordances}},
  author={Ahn, Michael and Brohan, Anthony and Brown, Noah and Chebotar, Yevgen and Cortes, Omar and David, Byron and Finn, Chelsea and Fu, Chuyuan and Gopalakrishnan, Keerthana and Hausman, Karol and others},
  journal={Conference on Robot Learning (CoRL)},
  year={2022}
}

@article{wilbert_colloseum,
  title={{THE COLOSSEUM: A Benchmark for Evaluating Generalization for Robotic Manipulation}},
  author={Pumacay, Wilbert and Singh, Ishika and Duan, Jiafei and Krishna, Ranjay and Thomason, Jesse and Fox, Dieter},
  journal={Robotics: Science and Systems (RSS)},
  year={2024}
}

@article{simpler_env,
  title={Evaluating Real-World Robot Manipulation Policies in Simulation},
          author={Xuanlin Li and Kyle Hsu and Jiayuan Gu and Karl Pertsch and Oier Mees and Homer Rich Walke and Chuyuan Fu and Ishikaa Lunawat and Isabel Sieh and Sean Kirmani and Sergey Levine and Jiajun Wu and Chelsea Finn and Hao Su and Quan Vuong and Ted Xiao},
  journal={Conference on Robot Learning (CoRL)},
  year={2024}
}

@article{droid,
    title   = {{DROID: A Large-Scale In-The-Wild Robot Manipulation Dataset}},
    author  = {Alexander Khazatsky and Karl Pertsch and Suraj Nair and Ashwin Balakrishna and Sudeep Dasari and Siddharth Karamcheti and Soroush Nasiriany and Mohan Kumar Srirama and Lawrence Yunliang Chen and Kirsty Ellis and Peter David Fagan and Joey Hejna and Masha Itkina and Marion Lepert and Yecheng Jason Ma and Patrick Tree Miller and Jimmy Wu and Suneel Belkhale and others},
    journal={Robotics: Science and Systems (RSS)},
    year    = {2024},
}

@article{xie2024decomposing,
  title={Decomposing the generalization gap in imitation learning for visual robotic manipulation},
  author={Xie, Annie and Lee, Lisa and Xiao, Ted and Finn, Chelsea},
  journal={International Conference on Robotics and Automation (ICRA)},
  year={2024},
}

@article{parekh2024investigating,
  title={{Investigating the Role of Instruction Variety and Task Difficulty in Robotic Manipulation Tasks}},
  author={Parekh, Amit and Vitsakis, Nikolas and Suglia, Alessandro and Konstas, Ioannis},
  journal={Conference on Empirical Methods in Natural Language Processing (EMNLP)},
  year={2024}
}

@article{kress2024robot,
  title={Robot Learning as an Empirical Science: Best Practices for Policy Evaluation},
  author={Kress-Gazit, Hadas and Hashimoto, Kunimatsu and Kuppuswamy, Naveen and Shah, Paarth and Horgan, Phoebe and Richardson, Gordon and Feng, Siyuan and Burchfiel, Benjamin},
  journal={arXiv},
  year={2024}
}

@article{anwar2024contrast,
  title={Contrast Sets for Evaluating Language-Guided Robot Policies},
  author={Anwar, Abrar and Gupta, Rohan and Thomason, Jesse},
  journal={Conference on Robot Learning (CoRL)},
  year={2024}
}

@article{bjorck2025gr00t,
  title={Gr00t n1: An open foundation model for generalist humanoid robots},
  author={Bjorck, Johan and Casta{\~n}eda, Fernando and Cherniadev, Nikita and Da, Xingye and Ding, Runyu and Fan, Linxi and Fang, Yu and Fox, Dieter and Hu, Fengyuan and Huang, Spencer and others},
  journal={arXiv preprint arXiv:2503.14734},
  year={2025}
}

@article{octo,
    title={Octo: An Open-Source Generalist Robot Policy},
    author = {{Octo Model Team} and Dibya Ghosh and Homer Walke and Karl Pertsch and Kevin Black and Oier Mees and Sudeep Dasari and Joey Hejna and Charles Xu and Jianlan Luo and Tobias Kreiman and {You Liang} Tan and Lawrence Yunliang Chen and Pannag Sanketi and Quan Vuong and Ted Xiao and Dorsa Sadigh and Chelsea Finn and Sergey Levine},
    journal = {Robotics: Science and Systems (RSS)},
    year = {2024},
}

@article{taxonomy2025arxiv,
  title={A Taxonomy for Evaluating Generalist Robot Policies},
  author={Gao, Jensen and Belkhale, Suneel and Dasari, Sudeep and Balakrishna, Ashwin and Shah, Dhruv and Sadigh, Dorsa},
  journal={IEEE Robotics and Automation Letters (RA-L)},
  year={2025}
}

@article{zhang2025rewind,
  title={ReWiND: Language-Guided Rewards Teach Robot Policies without New Demonstrations},
  author={Zhang, Jiahui and Luo, Yusen and Anwar, Abrar and Sontakke, Sumedh Anand and Lim, Joseph J and Thomason, Jesse and Biyik, Erdem and Zhang, Jesse},
  journal={Conference on Robot Learning (CoRL)},
  year={2025}
}

@inproceedings{brohan2022rt1,
      title={RT-1: Robotics Transformer for Real-World Control at Scale}, 
      author={Anthony Brohan and Noah Brown and Justice Carbajal and Yevgen Chebotar and Joseph Dabis and Chelsea Finn and Keerthana Gopalakrishnan and Karol Hausman and Alex Herzog and Jasmine Hsu and Julian Ibarz and Brian Ichter and others},
      year={2023},
      booktitle={Robotics: Science and Systems (RSS)},
}

@inproceedings{
kostrikov2022offline,
title={{Offline Reinforcement Learning with Implicit Q-Learning}},
author={Ilya Kostrikov and Ashvin Nair and Sergey Levine},
booktitle={International Conference on Learning Representations (ICLR)},
year={2022},
}

@article{nair2020awac,
  title={Awac: Accelerating online reinforcement learning with offline datasets},
  author={Nair, Ashvin and Gupta, Abhishek and Dalal, Murtaza and Levine, Sergey},
  journal={arXiv preprint arXiv:2006.09359},
  year={2020}
}

@article{kumar2019stabilizing,
  title={Stabilizing off-policy q-learning via bootstrapping error reduction},
  author={Kumar, Aviral and Fu, Justin and Soh, Matthew and Tucker, George and Levine, Sergey},
  journal={Neural Information Processing Systems (NeurIPS)},
  year={2019}
}

@inproceedings{fujimoto2019off,
  title={Off-policy deep reinforcement learning without exploration},
  author={Fujimoto, Scott and Meger, David and Precup, Doina},
  booktitle={International Conference on Machine Learning (ICML)},
  year={2019},
}

@article{nakamoto2023cal,
  title={Cal-ql: Calibrated offline rl pre-training for efficient online fine-tuning},
  author={Nakamoto, Mitsuhiko and Zhai, Simon and Singh, Anikait and Sobol Mark, Max and Ma, Yi and Finn, Chelsea and Kumar, Aviral and Levine, Sergey},
  journal={Neural Information Processing Systems (NeurIPS)},
  year={2023}
}

@inproceedings{kumar2020conservative,
  author       = {Aviral Kumar and Aurick Zhou and George Tucker and Sergey Levine},
  title        = {Conservative Q-Learning for Offline Reinforcement Learning},
  booktitle={Neural Information Processing Systems (NeurIPS)},
  year         = {2020},
}

@inproceedings{awr,
  author = {Peng, Xue Bin and Kumar, Aviral and Zhang, Grace and Levine, Sergey},
  title = {Advantage-Weighted Regression: Simple and Scalable Off-Policy Reinforcement Learning},
  booktitle = {arXiv preprint arXiv:1910.00177},
  year = {2019},
}

@inproceedings{open_x_embodiment_rt_x_2023,
title={Open {X-E}mbodiment: Robotic Learning Datasets and {RT-X} Models},
author = {Open X-Embodiment Collaboration and Abby O'Neill and Abdul Rehman and Abhinav Gupta and Abhiram Maddukuri and Abhishek Gupta and Abhishek Padalkar and Abraham Lee and Acorn Pooley and Agrim Gupta and Ajay Mandlekar and others},
booktitle={International Conference on Robotics and Automation (ICRA)},
year = {2024},
}

@article{nakamoto2024steering,
  author    = {Mitsuhiko Nakamoto and Oier Mees and Aviral Kumar and Sergey Levine},
  title     = {Steering Your Generalists: Improving Robotic Foundation Models via Value Guidance},
  journal   = {Conference on Robot Learning (CoRL)},
  year      = {2024},
}

@inproceedings{
kim2024openvla,
title={Open{VLA}: An Open-Source Vision-Language-Action Model},
author={Moo Jin Kim and Karl Pertsch and Siddharth Karamcheti and Ted Xiao and Ashwin Balakrishna and Suraj Nair and Rafael Rafailov and Ethan P Foster and Pannag R Sanketi and Quan Vuong and Thomas Kollar and Benjamin Burchfiel and Russ Tedrake and Dorsa Sadigh and Sergey Levine and Percy Liang and Chelsea Finn},
booktitle={Conference on Robot Learning (CoRL)},
year={2024},
}

@inproceedings{anwar2025efficient,
  title={Efficient Evaluation of Multi-Task Robot Policies With Active Experiment Selection},
  author={Anwar, Abrar and Gupta, Rohan and Merchant, Zain and Ghosh, Sayan and Neiswanger, Willie and Thomason, Jesse},
  booktitle={Conference on Robot Learning (CoRL)},
  year={2025}
}

@article{black2024pi0,
  title={$\pi_{0}$: A Vision-Language-Action Flow Model for General Robot Control},
  author={Kevin Black and Noah Brown and Danny Driess and Adnan Esmail and Michael Equi and Chelsea Finn and others},

    year={2024}
}

@article{physicalintelligence2025pi0.5,
  title={$\pi_{0.5}$: A Vision-Language-Action Model with Open-World Generalization},
  author={{Physical Intelligence} and others},
  journal={arXiv preprint arXiv:2504.16054},
  year={2025}
}

@article{mitra2025roboticsteering,
  title={Mechanistic Finetuning of Vision-Language-Action Models via Few-Shot Demonstrations},
  author={Mitra, Chancharik and Luo, Yusen and Saravanan, Raj and Niu, Dantong and Pai, Anirudh and Thomason, Jesse and Darrell, Trevor and Anwar, Abrar and Ramanan, Deva and Herzig, Roei},
  journal={European Conference on Computer Vision (ECCV)},
  year={2026}
}

@article{liang2026robometer,
  title={Robometer: Scaling general-purpose robotic reward models via trajectory comparisons},
  author={Liang, Anthony and Korkmaz, Yigit and Zhang, Jiahui and Hwang, Minyoung and Anwar, Abrar and Kaushik, Sidhant and Shah, Aditya and Huang, Alex S and Zettlemoyer, Luke and Fox, Dieter and others},
  journal={Robotics: Science and Systems (RSS)},
  year={2026}
}

@inproceedings{pertsch2025fast,
  title     = {{FAST}: Efficient Action Tokenization for Vision-Language-Action Models},
  author    = {Pertsch, Karl and Stachowicz, Kyle and Ichter, Brian and Driess, Danny and Nair, Suraj and Vuong, Quan and Mees, Oier and Finn, Chelsea and Levine, Sergey},
  booktitle = {Robotics: Science and Systems (RSS)},
  year      = {2025}
}

@inproceedings{kim2025openvlaoft,
  title     = {Fine-Tuning Vision-Language-Action Models: Optimizing Speed and Success},
  author    = {Kim, Moo Jin and Finn, Chelsea and Liang, Percy},
  booktitle = {Robotics: Science and Systems (RSS)},
  year      = {2025}
}


\end{document}